%% file: main.tex
\documentclass[11pt,a4paper]{article}
\usepackage[hyperref]{naaclhlt2018}
\usepackage{times}
\usepackage{latexsym}
\usepackage{verbatim}

\usepackage{url}
\usepackage{graphicx}

\usepackage{tabularx}
\usepackage{amsmath}
\usepackage{amssymb}
\usepackage{amsfonts}
\usepackage{multirow}
\usepackage{booktabs}
\usepackage{arydshln}
\usepackage[bb=fourier]{mathalfa}
\usepackage{xcolor}

\usepackage{caption}

\DeclareCaptionFont{10pt}{\fontsize{10pt}{12pt}\selectfont}
\DeclareMathOperator*{\argmax}{argmax}

\aclfinalcopy
\def\aclpaperid{440}

\title{Unified Pragmatic Models for Generating and Following Instructions
}

\author{Daniel Fried ~~ Jacob Andreas ~~ Dan Klein \\
 Computer Science Division \\
 University of California, Berkeley\\
 {\tt \{dfried,jda,klein\}@cs.berkeley.edu}}

\date{}

\begin{document}

\maketitle

\begin{abstract}
We show that explicit pragmatic inference aids in correctly generating and following natural language instructions for complex, sequential tasks.  Our pragmatics-enabled models reason about why speakers produce certain instructions, and about how listeners will react upon hearing them.  Like previous pragmatic models, we use learned base listener and speaker models to build a \emph{pragmatic speaker} that uses the base listener to simulate the interpretation of candidate descriptions, and a \emph{pragmatic listener} that reasons counterfactually about alternative descriptions. We extend these models to tasks with sequential structure.  Evaluation of language generation and interpretation shows that pragmatic inference improves state-of-the-art listener models (at correctly interpreting human instructions) and speaker models (at producing instructions correctly interpreted by humans) in diverse settings.
\end{abstract}

\section{Introduction}

How should speakers and listeners reason about each other when they communicate? A core insight of computational pragmatics  is that speaker and listener agents operate within a cooperative game-theoretic context, and that each agent benefits from
reasoning about others' intents and actions within that context.
Pragmatic inference has been studied by a long
line of work in linguistics, natural language processing, and cognitive science.
In this paper, we present a technique for layering explicit pragmatic inference on top of models for complex, sequential instruction-following and instruction-generation tasks.  We investigate a range of current data sets for both tasks, showing that pragmatic behavior arises naturally from this inference procedure, and gives rise to state-of-the-art results in a variety of domains.

\begin{figure}
\centering
\strut \\
\vspace{-1.5em}
\strut
\raisebox{4.0em}{(a)} \includegraphics[width=0.85\columnwidth,trim=0 5.2in 4.5in 0,clip]{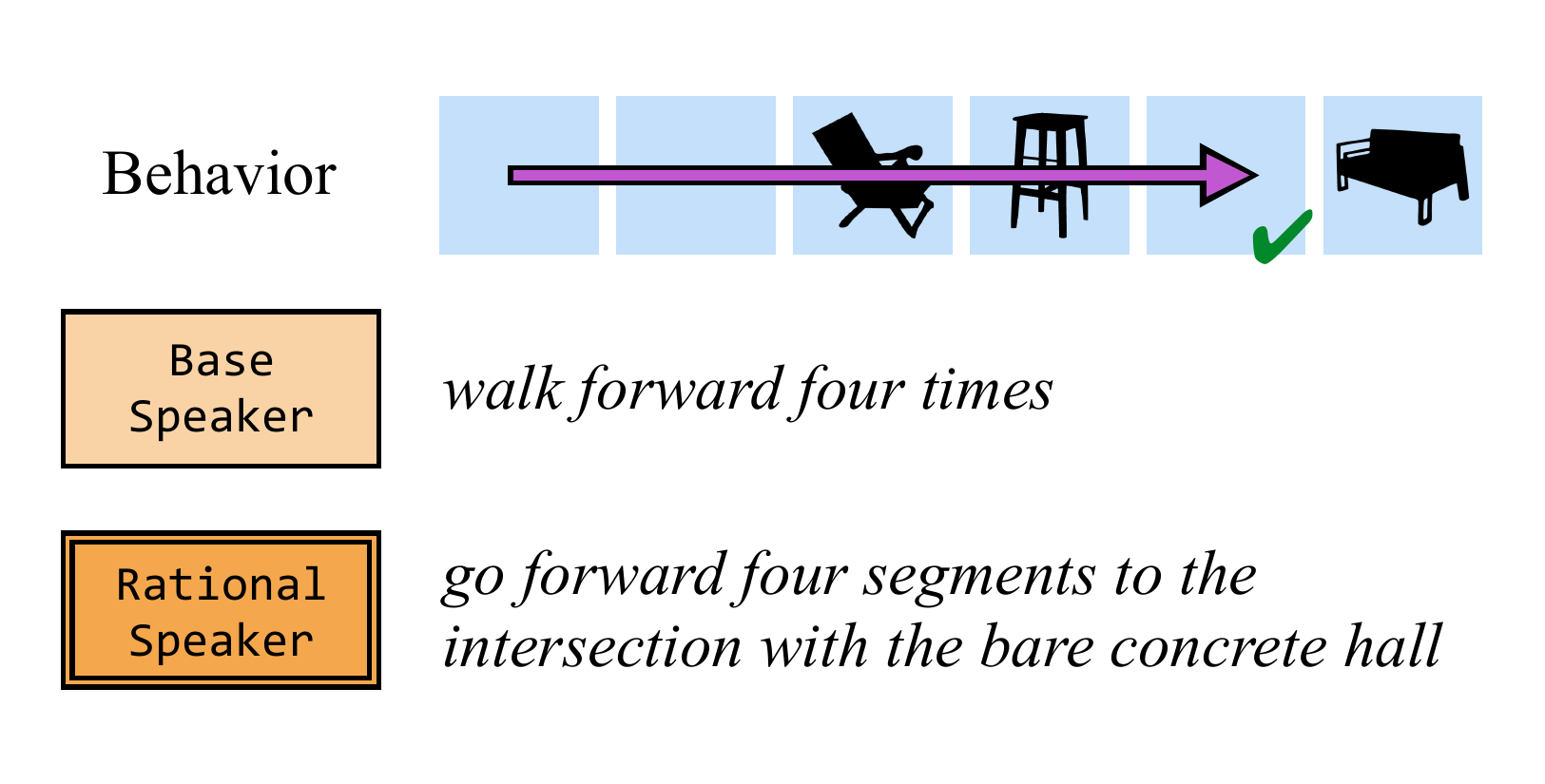} \\[-.5em]
\hrule
\raisebox{4.0em}{(b)} \includegraphics[width=0.85\columnwidth,trim=0 5.2in 4.5in 0,clip]{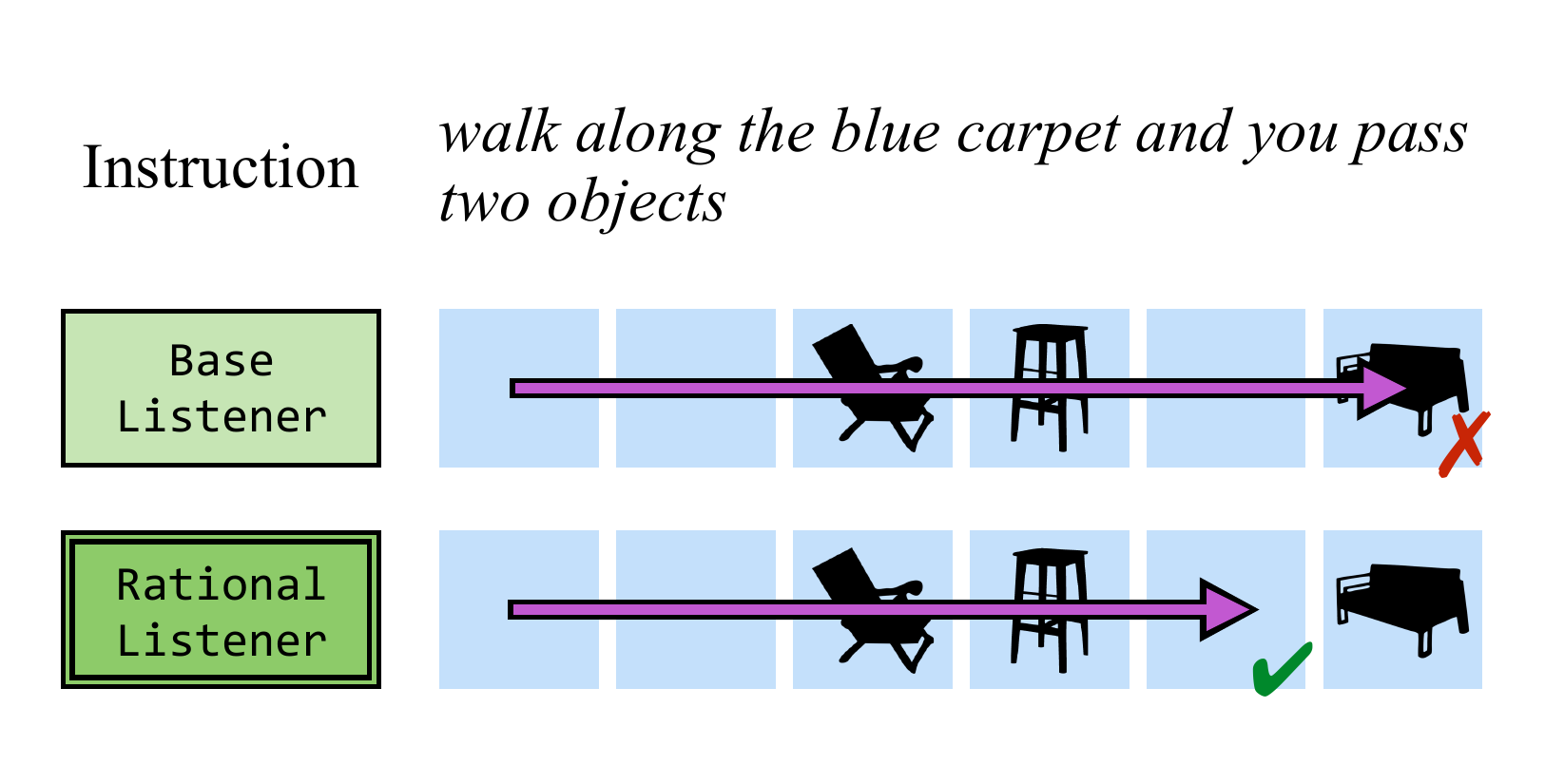} \\[-.6em]
\caption{Real samples for the SAIL navigation environments, comparing \emph{base} models, without explicit pragmatic inference, to the \emph{rational} pragmatic inference procedure.
(a) The rational speaker, which reasons about listener behavior, generates 
instructions which in this case are more robust to uncertainty about the listener's initial orientation. 
(b) The base listener moves to an unintended position (even though it
correctly \emph{passes two objects}). The rational listener, which reasons about the speaker, infers that a route ending at the sofa would have been described differently, and stops earlier. 
}
\label{fig:teaser}
\vspace{-1em}
\end{figure}

Consider the example shown in
\autoref{fig:teaser}a, in which a speaker agent must describe a route to a
target position in a hallway. A conventional learned instruction-generating
model produces a truthful description of the route (\emph{walk forward four times}).
But the pragmatic speaker in this paper, which is capable of
reasoning about the listener, chooses to also include additional information
(\emph{the intersection with the bare concrete hall}), to reduce potential
ambiguity and increase the odds that the listener reaches the correct
destination. 

This same reasoning procedure also allows a listener agent to overcome ambiguity
in instructions by reasoning counterfactually about the speaker
(\autoref{fig:teaser}b). Given the command \emph{walk along the blue carpet and
you pass two objects}, a conventional learned instruction-following model 
is willing to consider all paths that pass two objects, and ultimately arrives
at an unintended final position. But a pragmatic listener that reasons about
the speaker can infer that the long path would have been more easily described
as \emph{go to the sofa}, and thus that the shorter path is probably intended.
In these two examples, which are produced by the system we describe in this
paper, a unified reasoning process (choose the output sequence which is most
preferred by an embedded model of the other agent) produces pragmatic behavior
for both speakers and listeners. 

The application of models with explicit pragmatic reasoning abilities has so far
been largely restricted to simple reference games, in which the
listener's only task is to select the right item from among a small set of
candidate referents given a single short utterance from the speaker. But as the
example shows, there are real-world instruction following and generation tasks
with rich action spaces that might also benefit from pragmatic
modeling. 
Moreover, approaches that learn to map directly between human-annotated instructions and action sequences are ultimately limited by the effectiveness of the humans themselves. The promise of pragmatic modeling is that we can use these same annotations to build a model with a different (and perhaps even better) mechanism for interpreting and generating instructions.

The primary contribution of this work is to show how existing models of pragmatic
reasoning can be extended to support instruction following and generation for challenging, multi-step, interactive tasks.
Our experimental evaluation focuses on four instruction-following domains which
have been studied using both semantic parsers and attentional neural models.  We
investigate the interrelated tasks of instruction following and instruction
generation, and show that incorporating an explicit model of pragmatics 
helps in both cases. 
Reasoning about the human listener allows a speaker model to produce instructions
that are easier for humans to interpret correctly in all domains (with absolute
gains in accuracy ranging from 12\% to 46\%). 
Similarly, reasoning about the human speaker improves the accuracy of the
listener models in interpreting instructions in most domains (with gains in
accuracy of up to 10\%).  In all cases, the resulting systems are competitive
with, and in many cases exceed, results from past state-of-the-art systems for
these tasks.\footnote{Source code is available at \url{http://github.com/dpfried/pragmatic-instructions}}

\section{Problem Formulation}
\label{sec:tasks}
Consider the instruction following and instruction generation tasks shown in \autoref{fig:teaser}, where an agent must produce or interpret instructions about a structured world context (e.g.\ \emph{walk along the blue carpet and you pass two objects}). 

In the \textbf{instruction following} task, a listener agent begins in a world state (in \autoref{fig:teaser} an initial map location and orientation). 
The agent is then tasked with following a sequence of direction sentences $d_1 \ldots d_K$ produced by humans. 
At each time $t$ the agent receives a percept $y_t$, which is a feature-based representation of the current world state, and chooses an action $a_t$ (e.g.\ move forward, or turn). The agent succeeds if it is able to reach the correct final state described by the directions.

In the \textbf{instruction generation} task, the agent receives a sequence of actions $a_1, \cdots a_T$ along with the world state $y_1, \cdots y_T$ at each action, and must generate a sequence of direction sentences $d_1, \ldots d_K$ describing the actions. The agent succeeds if a human listener is able to correctly follow those directions to the intended final state.

We evaluate models for both tasks in four domains. The first domain is the SAIL corpus of virtual environments and navigational directions~\cite{MacMahon06SAIL,Chen11Navigation}, where an agent navigates through a two-dimensional grid of hallways with patterned walls and floors and a discrete set of objects (\autoref{fig:teaser} shows a portion of one of these hallways).

In the three SCONE domains~\cite{long2016simpler}, the world contains a number of objects with various properties, such as colored beakers which an agent can combine, drain, and mix. Instructions describe how these objects should be manipulated. These domains were designed to elicit instructions with a variety of context-dependent language phenomena, including ellipsis and coreference~\cite{long2016simpler} which we might expect a model of pragmatics to help resolve \cite{Potts11Handbook}.

\section{Related Work}
The approach in this paper builds upon long lines of work in pragmatic modeling, instruction following, and instruction generation.

\paragraph{Pragmatics}

Our approach to pragmatics \cite{Grice75} belongs to a general category of rational speech acts
models \cite{frank2012predicting}, in which the interaction between speakers
and listeners is modeled as a probabilistic process with Bayesian actors~\cite{goodman2013knowledge}. Alternative
formulations (e.g.\ with best-response rather than probabilistic dynamics) are also
possible \cite{Golland10Game}. Inference in these models is challenging even when the space of listener actions is extremely simple \cite{Smith13BayesianPragmatics}, and one of our goals in the present work is to show how this inference problem can be solved even in much richer action spaces than previously considered in computational pragmatics.
This family of pragmatic models captures a number of important linguistic phenomena,
especially those involving conversational implicature \cite{Monroe15RationalSpeech}; we  note that many other topics studied under the broad heading of ``pragmatics,'' including presupposition and indexicality, require different machinery.

\newcite{williams2015going} use pragmatic reasoning with weighted inference rules to resolve ambiguity and generate clarification requests in a human-robot dialog task. Other recent work on pragmatic models focuses on the referring expression generation or ``contrastive captioning'' task introduced by \newcite{Kazemzadeh14ReferIt}. In this family are approaches that model the listener at training time \cite{Mao15Generation}, at evaluation time \cite{Andreas16Pragmatics,monroe2017colors,vedantam2017context,su2017reasoning} or both \cite{yu2017refexpr,luo2017comprehension}. 

Other conditional sequence rescoring models that are structurally similar but motivated by concerns other than pragmatics include \newcite{li2015diversity} and \newcite{yu2017neural}. \newcite{lewis2017deal} perform a similar inference procedure for a competitive negotiation task. The language learning model of \newcite{wang2016learning} also features a structured output space and uses pragmatics to improve online predictions for a semantic parsing model. Our approach in this paper performs both generation and interpretation, and investigates both structured and unstructured output representations.

\begin{figure*}

	\centering
    \hfill
    \includegraphics[height=1.6in,trim=0in 3.4in 5.65in 0in,clip]{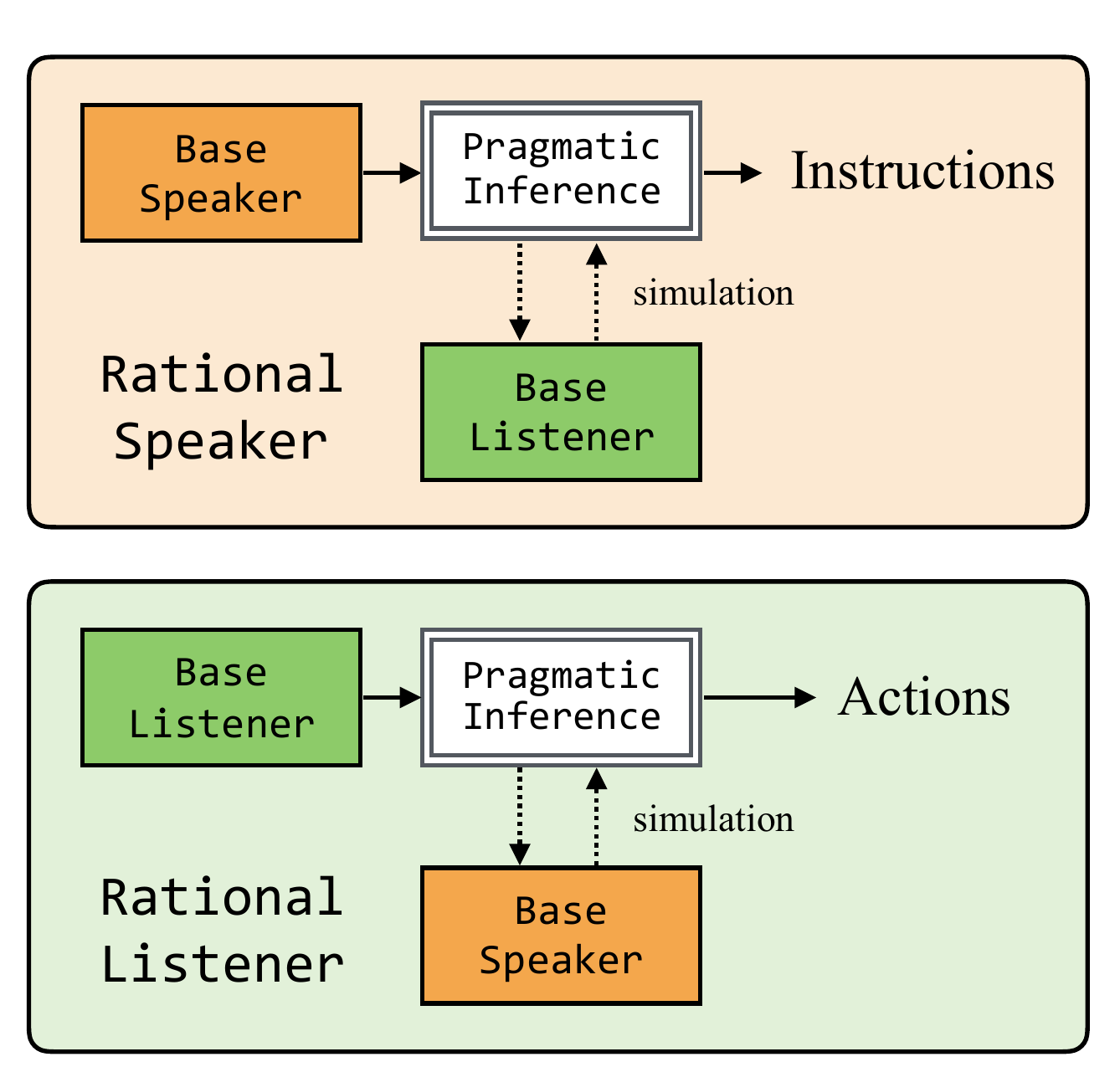}
    \hfill
    \includegraphics[height=1.6in,trim=0in 4.4in 1.2in 0in,clip]{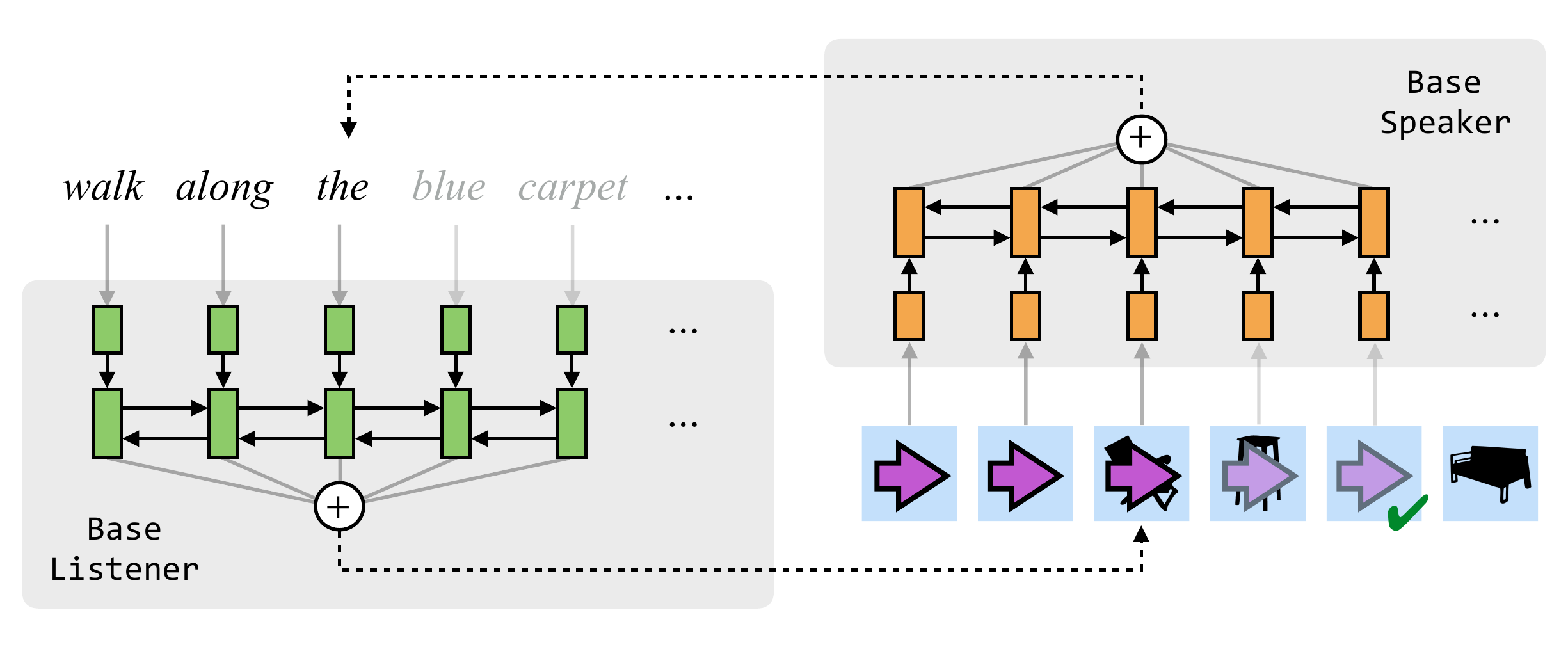} 
    \hfill
    \strut
    \\
    \footnotesize \strut \hspace{0.7in} (a) \hspace{2.8in} (b) \hspace{1.7in} \strut \\[-1em]
    \caption{
    (a) Rational pragmatic models embed base listeners and speakers. Potential candidate sequences are drawn from one base model, and then the other scores each candidate to simulate whether it produces the desired pragmatic behavior.
    (b) The base listener and speaker are neural sequence-to-sequence models which are largely symmetric to each other. Each produces a representation of its input sequence (a description, for the listener; actions with associated environmental percepts, for the listener) using an LSTM encoder. The output sequence is generated by an LSTM decoder attending to the input.}
    \label{fig:models}
    \vspace{-1em}
\end{figure*}

\paragraph{Instruction following}

Work on instruction following tasks includes models that parse commands
into structured representations processed by a rich execution model 
\cite{Tellex11Commands,Chen12Online,Artzi13Navigation,guu2017language}, and models that map directly
from instructions to a policy over primitive actions \cite{Branavan09PG}, possibly
mediated by an intermediate alignment or attention variable 
\cite{Andreas15Instructions,Mei16Instructions}. 
We use a model similar to \citet{Mei16Instructions} as our base listener in this paper, evaluating on the SAIL navigation task \cite{MacMahon06SAIL} as they did, as well as the SCONE context-dependent execution domains~\cite{long2016simpler}.

\paragraph{Instruction generation}
Previous work has also investigated the instruction generation task, in particular for navigational directions. The GIVE shared tasks  \cite{give1,give2,give2_5} have produced a large number of interactive direction-giving systems, both rule-based and learned. The work most immediately related to the generation task in this paper is that of \newcite{daniele2016navigational}, which also focuses on the SAIL dataset but requires substantial additional structured annotation for training, while both our base and pragmatic speaker models learn directly from strings and action sequences.  

Older work has studied the properties of effective human strategies for
generating navigational directions \cite{Anderson91MapTask}. Instructions of
this kind can be used to extract templates for generation
\cite{look2008cognitively,dale2005using}, while here we focus on the more
challenging problem of learning to generate new instructions from scratch. Like
our pragmatic speaker model, \newcite{goeddel2012dart} also reason about
listener behavior when generating navigational instructions, but rely on
rule-based models for interpretation.

\section{Pragmatic inference procedure}
\label{sec:rational_models}

As a foundation for pragmatic inference, we assume that we have \emph{base} listener and speaker models to map directions to actions and vice-versa. (Our notation for referring to models is adapted from \citet{bergen16rsa}.)
The base listener, $L_0$, produces a probability distribution over sequences of actions, conditioned on a representation of the directions and environment as seen before each action: 
$P_{L_0}(a_{1:T} | d_{1:K}, y_{1:T})$.
Similarly, the base speaker, $S_0$, defines a distribution over possible descriptions conditioned on a representation of the actions and environment:
$P_{S_0}(d_{1:K}|a_{1:T},y_{1:T})$.

Our pragmatic inference procedure requires these base models to produce candidate outputs from a given input (actions from descriptions, for the listener; descriptions from actions, for the speaker), and calculate the probability of a fixed output given an input, but is otherwise agnostic to the form of the models.

We use standard sequence-to-sequence models with attention for both the base listener and speaker (described in \autoref{sec:base_details}). Our models use segmented action sequences, with one segment (sub-sequence of actions) aligned with each description sentence $d_{j}$, for all $j \in \{1 \ldots K\}$. This segmentation is either given as part of the training and testing data (in the instruction following task for the SAIL domain, and in both tasks for the SCONE domain, where each sentence corresponds to a single action), or is predicted by a separate segmentation model (in the generation task for the SAIL domain), see \autoref{sec:base_details}.

\subsection{Models}

Using these base models as self-contained modules, we derive a \emph{rational speaker} and \emph{rational listener} that perform inference using embedded instances of these base models (\autoref{fig:models}a).
When describing an action sequence, a rational speaker $S_1$ chooses a description that has a high chance of causing the listener modeled by $L_0$ to follow the given actions:
\begin{equation}
\label{eq:rational_speaker}
     S_1(a_{1:T}) = \argmax_{d_{1:K}} P_{L_0}(a_{1:T} | d_{1:K}, y_{1:T})
\end{equation}
(noting that, in all settings we explore here, the percepts $y_{1:T}$ are completely determined by the actions $a_{1:T}$).
Conversely, a rational listener $L_1$ follows a description by choosing an action sequence which has high probability of having caused the speaker, modeled by $S_0$, to produce the description:
\begin{equation}
\label{eq:rational_listener}
    \hspace{-0.05em} L_1(d_{1:K}) = \argmax_{a_{1:T}} P_{S_0}(d_{1:K} | a_{1:T}, y_{1:T})
\end{equation}

These optimization problems are intractable to solve for general base listener and speaker agents, including the sequence-to-sequence models we use, as they involve choosing an input (from a combinatorially large space of possible sequences) to maximize the probability of a fixed output sequence. We instead follow a simple approximate inference procedure, detailed in Section \ref{sec:inference}.

We consider also incorporating the scores of the base model used to produce the candidates. For the case of the speaker, we define a \emph{combined rational speaker}, denoted $S_0 \cdot S_1$, that selects the candidate that maximizes a weighted product of probabilities under both the base listener and the base speaker:
\begin{align}
    \label{eq:combined_speaker}
    \argmax_{d_{1:K}} & P_{L_0} (a_{1:T}|d_{1:K}, y_{1:T})^{\lambda} \nonumber \\[-0.5em]
    &\times~P_{S_0}(d_{1:K} | a_{1:T}, y_{1:T})^{1-\lambda}
\end{align}
for a fixed interpolation hyperparameter $\lambda \in [0,1]$. There are several motivations for this combination with the base speaker score. First, as argued by \citet{monroe2017colors}, we would expect varying degrees of base and reasoned interpretation in human speech acts.
Second, we want the descriptions produced by the model to be fluent descriptions of the actions. Since the base models are trained discriminatively, maximizing the probability of an output sequence for a fixed input sequence, their scoring behaviors for fixed outputs paired with inputs dissimilar to those seen in the training set may be poorly calibrated 
(for example when conditioning on ungrammatical descriptions).
Incorporating the scores of the base model used to produce the candidates 
aims to prevent this behavior. 

To define rational listeners, we use the symmetric formulation: first, draw candidate action sequences from $L_0$. For $L_1$, choose the actions that achieve the highest probability under $S_0$; and for the combination model $L_0 \cdot L_1$ choose the actions with the highest weighted combination of $S_0$ and $L_0$ (paralleling equation~\ref{eq:combined_speaker}).

\subsection{Inference}
\label{sec:inference}

As in past work \cite{Smith13BayesianPragmatics,Andreas16Pragmatics,monroe2017colors}, we approximate the optimization problems in equations~\ref{eq:rational_speaker}, \ref{eq:rational_listener}, and \ref{eq:combined_speaker}: use the base models to generate 
candidates, and rescore them to find ones that are likely to produce the desired 
behavior. 

In the case of the rational speaker $S_1$, we use the base speaker $S_0$ to produce a set of $n$ candidate descriptions $w^{(1)}_{1:K_1} \ldots w^{(n)}_{1:K_n}$ for the sequences $a_{1:T}, y_{1:T}$, using beam search.
We then find the score of each description under $P_{L_0}$ (using it as the input sequence for the observed output actions we want the rational speaker to describe), or a weighted combination of $P_{L_0}$ and the original candidate score $P_{S_0}$, and choose the description $w_{1:K_j}^{(j)}$ with the largest score, approximately solving the maximizations in equations~\ref{eq:rational_speaker} or \ref{eq:combined_speaker}, respectively. 
We perform a symmetric procedure for the rational listener: produce action sequence candidates from the base listener, and rescore them using the base speaker.\footnote{We use ensembles of models for the base listener and speaker (\autoref{sec:training}), and to obtain candidates that are high-scoring under the combination of models in the ensemble, we perform standard beam search using all models in lock-step. At every timestep of the beam search, each possible extension of an output sequence is scored using the product of the extension's conditional probabilities across all models in the ensemble.}

As the rational speaker must produce long output sequences (with multiple sentences), we interleave the speaker and listener in inference, determining each output sentence sequentially. From a list of candidate direction sentences from the base speaker for the current subsequence of actions, we choose the top-scoring direction under the listener model (which may also condition on the directions which have been output previously), and then move on to the next subsequence of actions.\footnote{We also experimented with sampling from the base models to produce these candidate lists, as was done in previous work \cite{Andreas16Pragmatics,monroe2017colors}. In early experiments, however, we found better performance with beam search in the rational models for all tasks.
}

\section{Base model details}
\label{sec:base_details}
Given this framework, all that remains is to describe the base models $L_0$ and $S_0$.
We implement these as sequence-to-sequence models that map directions to actions (for the listener) or actions to directions (for the speaker), additionally conditioning on the world state at each timestep.

\subsection{Base listener}
\label{sec:base_listener}
Our base listener model, $L_0$, predicts action sequences conditioned on an encoded representation of the directions and the current world state.
In the SAIL domain, this is the model of 
\newcite{Mei16Instructions}  (illustrated in green in \autoref{fig:models}b for a single sentence and its associated actions), see ``domain specifics'' below.

\paragraph{Encoder}
Each direction sentence is encoded separately with a bidirectional LSTM \citep{hochreiter1997long}; the LSTM's hidden states are reset for each sentence. We obtain a representation $h^e_k$ for the $k$th word in the current sentence by concatenating an embedding for the word with its forward and backward LSTM outputs.

\paragraph{Decoder}
We generate actions incrementally using an LSTM decoder with monotonic alignment between the direction sentences and subsequences of actions; at each timestep the decoder predicts the next action for the current sentence $w_{1:M}$ (including choosing to shift to the next sentence).
The decoder takes as input at timestep $t$ the current world state, $y_t$ and a representation $z_t$ of the current sentence, updates the decoder state $h^d$, 
and outputs a distribution over possible actions:
\begin{align}
  h^d_t &= \text{LSTM}_{d}(h^d_{t-1}, [W_y y_t, z_t]) \nonumber \\
  q_t &= W_o (W_y y_t + W_h h^d_t + W_z z_t)  \nonumber \\
  p(a_t &\mid a_{1:t-1}, y_{1:t}, w_{1:M}) \propto \exp(q_t) \nonumber
\end{align}
where all weight matrices $W$ are learned parameters.
The sentence representation $z_t$ is produced using an attention mechanism ~\cite{bahdanau2014neural} over the representation vectors $h^e_1 \ldots h^e_M$ for words in the current sentence:
\begin{align}
  \alpha_{t,k} &\propto \exp(v \cdot \tanh(W_d h^d_{t-1} + W_e h_{k}^e)) \nonumber \\
  z_t &= \sum_{k=1}^M \alpha_{t,k} h_k^e \nonumber
\end{align}
where the attention weights $\alpha_{t,k}$ are normalized to sum to one across positions $k$ in the input, and weight matrices $W$ and vector $v$ are learned.

\paragraph{Domain specifics}
For SAIL, we use the alignments between sentences and route segments annotated by \newcite{Chen11Navigation}, which were also used in previous work \cite{Artzi13Navigation,Artzi14Compact,Mei16Instructions}.
Following \newcite{Mei16Instructions}, we reset the decoder's hidden state for each sentence. 

In the SCONE domains, which have a larger space of possible outputs 
than SAIL, we extend the decoder by: (i) decomposing each action into an action type and arguments for it, (ii) using separate attention mechanisms for types and arguments  and (iii) using state-dependent action embeddings. See Appendix~\ref{sec:appendix} in the supplemental material for details. The SCONE domains are constructed so that each sentence corresponds to a single (non-decomposed) action; this provides our segmentation of the action sequence.

\subsection{Base speaker}
\label{sec:base_speaker}
While previous work \cite{daniele2016navigational} has relied on more structured approaches, we construct our base speaker model $S_0$ using largely the same sequence-to-sequence
machinery as above.
$S_0$ (illustrated in orange in \autoref{fig:models}b) encodes a sequence of actions and world states, and then uses a decoder to output a description.

\paragraph{Encoder}
We  encode the sequence of vector embeddings for the actions $a_t$ and world states $y_t$ using a bidirectional LSTM. Similar to the base listener's encoder, we then obtain a representation $h_t^e$ for timestep $t$ by concatenating $a_t$ and $y_t$ with the LSTM outputs at that position.

\paragraph{Decoder}
As in the listener, we use an LSTM decoder with monotonic alignment between direction sentences and subsequences of actions, and attention over the subsequences of actions. The decoder takes as input at position $k$ an embedding for the previously generated word $w_{k-1}$ and a representation $z_k$ of the current subsequence of actions and world states, and produces a distribution over words (including ending the description for the current subsequence and advancing to the next). The decoder's output distribution is produced by:
\begin{align*}
  h^d_k &= \text{LSTM}_{d}(h^d_{k-1}, [w_{k-1}, z_k]) \nonumber \\
  q_k &= W_h h^d_k + W_z z_k \nonumber \\
  p(w_k &\mid w_{1:k-1}, a_{1:T}, y_{1:T}) \propto \exp(q_k)
\end{align*}
where all weight matrices $W$ are learned parameters.\footnote{All parameters are distinct from those used in the base listener; the listener and speaker are trained separately.}
As in the base listener, the input representation $z_k$ is produced by attending to the vectors  $h^e_1 \ldots h^e_T$ encoding the input sequence (here, encoding the subsequence of actions and world states to be described):
\begin{align*}
  \alpha_{k,t} &\propto \exp(v \cdot \tanh(W_d h^d_{k-1} + W_e h_{t}^e)) \nonumber \\
  z_k &= \sum_{t=1}^T \alpha_{k,t}~h_t^e \nonumber
\end{align*}
The decoder's LSTM state is reset at the beginning of each sentence.

\paragraph{Domain specifics}
In SAIL, for comparison to the generation system of \newcite{daniele2016navigational} which did not use segmented routes, we train a route segmenter for use at test time. We also represent routes using a collapsed representation of action sequences.
In the SCONE domains, we (i) use the same context-dependent action embeddings used in the listener, and (ii) don't require an attention mechanism, since only a single action is used to produce a given sentence  within the sequence of direction sentences. 
See Appendix~\ref{sec:appendix} for more details.

\subsection{Training}
\label{sec:training}
The base listener and speaker models are trained independently to maximize the conditional
likelihoods of the actions--directions pairs in the training sets.
See Appendix~\ref{sec:appendix} for details on the optimization, LSTM variant, and hyperparameters.

We use ensembles for the base listener $L_0$ and base speaker $S_0$, where each ensemble consists of 10 models trained from separate random parameter initializations. This follows the experimental setup of \citet{Mei16Instructions} for the SAIL base listener. 

\section{Experiments}
\label{sec:experiments}
We evaluate speaker and listener agents on both the instruction following and instruction generation tasks in the SAIL domain and three SCONE domains (\autoref{sec:tasks}). 
For all domains, we compare the rational listener and speaker against the base listener and speaker, as well as against past state-of-the-art results for each task and domain. Finally, we examine pragmatic inference from a model combination perspective, comparing the pragmatic reranking procedure to ensembles of a larger number of base speakers or listeners.

For all experiments, we use beam search both to generate candidate lists for the rational systems (section~\ref{sec:inference}) and to generate the base model's output. We fix the beam size $n$ to be the same in both the base and rational systems, using $n=20$ for the speakers and $n=40$ for the listeners. We tune the weight $\lambda$ in the combined rational agents ($L_0\cdot L_1$ or $S_0 \cdot S_1$) to maximize accuracy (for listener models) or BLEU (for speaker models) on each domain's development data.

\subsection{Instruction following}
We evaluate our listener models by their accuracy in carrying out human instructions: whether the systems were able to reach the final world state which the human was tasked with guiding them to. 

\paragraph{SAIL}
\begin{table}[t]
\centering
\footnotesize
\begin{tabular}{rcccc}
  \toprule
& \multicolumn{2}{c}{Single-sentence} & \multicolumn{2}{c}{Multi-sentence} \\
listener & Rel & Abs & Rel & Abs \\
\cmidrule(lr){1-1} \cmidrule(lr){2-3} \cmidrule(lr){4-5}
\multirow{2}{*}{past work} & 69.98 & 65.28 & 26.07 & \phantom{+}35.44\\
                           & (MBW) & (AZ) & (MBW) & \phantom{+}(ADP) \\
\hdashline \\[-0.8em]
$L_0$ & 68.40 & 59.62 & 24.79 & \phantom{+}13.53 \\
$L_0 \cdot L_1$ & \bf 71.64 & 64.38 & \bf 34.05 & \phantom{+}24.50 \\
\emph{accuracy gain} & +3.24 & +4.76 & +9.26 & +10.97 \\
\bottomrule
\end{tabular}
\caption{\label{tbl:follower_results} Instruction-following results
on the SAIL dataset. The table shows cross-validation test accuracy for the base listener ($L_0$) and pragmatic listeners ($L_0 \cdot L_1$), along with the gain  given by pragmatics. We report results for the single- and multi-sentence conditions, under the relative and absolute starting conditions\textsuperscript{\ref{note:orientation}}, comparing to the best-performing prior work by \citet{Artzi13Navigation} (AZ), \citet{Artzi14Compact} (ADP), and \citet{Mei16Instructions} (MBW). Bold numbers show new state-of-the-art results.}
\vspace{-0.75em}
\end{table}

We follow standard cross-validation evaluation for the instruction following task on the SAIL dataset \cite{Artzi13Navigation,Artzi14Compact,Mei16Instructions}.\footnote{\label{note:orientation}Past work has differed in the handling of undetermined orientations in the routes, which occur in the first state for multi-sentence routes and the first segment of their corresponding single-sentence routes. For comparison to both types of past work, we train and evaluate listeners in two settings: \emph{Abs}, which sets these undetermined starting orientations to be a fixed absolute orientation, and \emph{Rel}, where an undetermined starting orientation is set to be a 90 degree rotation from the next state in the true route.}
\autoref{tbl:follower_results} shows improvements over the base listener $L_0$ when using the rational listener $L_0 \cdot L_1$ in the single- and multi-sentence settings. We also report the best accuracies from past work. 
We see that the largest relative gains come in the multi-sentence setting, where handling ambiguity is potentially more important to avoid compounding errors.
The rational model improves on the published results of \citet{Mei16Instructions}, and while it is still below the systems of \citet{Artzi13Navigation} and \citet{Artzi14Compact}, which use additional supervision in the form of hand-annotated seed lexicons and logical domain representations, it approaches their results in the single-sentence setting.

\paragraph{SCONE}

\begin{table}[t]

\centering
\footnotesize
\begin{tabular}{rccc}
  \toprule
  listener & Alchemy & Scene & Tangrams \\
  \midrule
  GPLL & 52.9 & 46.2 & 37.3 \\
  \hdashline \\[-0.8em]
  $L_0$ & 69.7 & 70.9 & 69.6 \\
  $L_0 \cdot L_1$ & 72.0 & 72.7 & 69.6 \\
  \emph{accuracy gain} & +2.3 & +1.8 & +0.0 \\
  \bottomrule
\end{tabular}
\caption{\label{tbl:scone_follower_results}Instruction-following results in the SCONE domains. The table shows accuracy on the test set. For reference, we also show prior results from \citet{guu2017language} (GPLL), although our models use more supervision at training time.}
\end{table}

\begin{figure}[t]
\footnotesize
\begin{tabular}{ll}
\multicolumn{2}{l}{
\texttt{a red guy appears on the far left}
} \\
\multicolumn{2}{l}{
\texttt{then to orange's other side}
} \\
\multicolumn{2}{l}{} \\
base listener, $L_0$ & rational listener, $L_0 \cdot L_1$ \\[0.5em]
\includegraphics[width=0.45\linewidth]{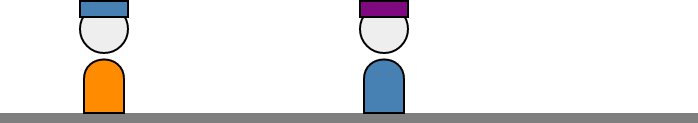} &  \includegraphics[width=0.45\linewidth]{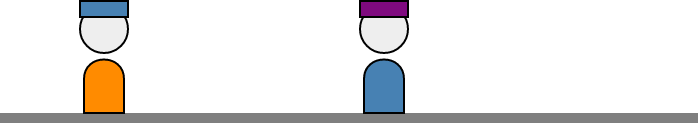} \\
\includegraphics[width=0.45\linewidth]{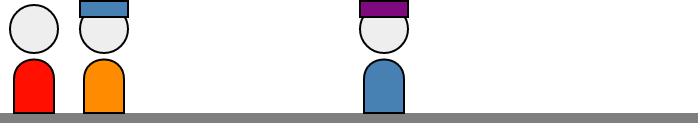} &  \includegraphics[width=0.45\linewidth]{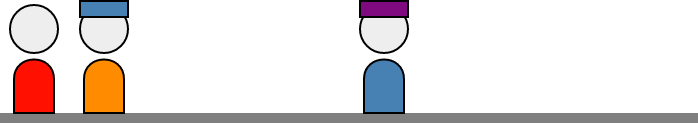} \\
\includegraphics[width=0.45\linewidth]{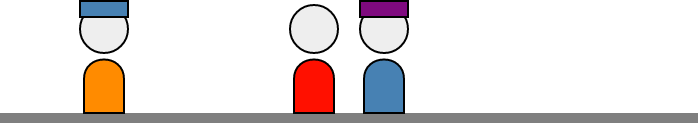} &  \includegraphics[width=0.45\linewidth]{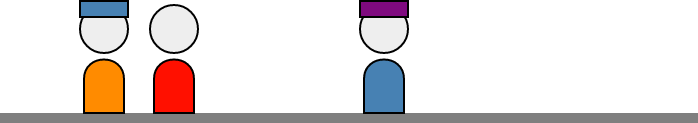} \\
\end{tabular}
\caption{\label{fig:alchemy_listener}Action traces produced for a partial instruction sequence (two instructions out of five) in the Scene domain. The base listener moves the red figure to a position that is a marginal, but valid, interpretation of the directions. 
The rational listener correctly produces the action sequence the directions were intended to describe.}
\vspace{-1em}
\end{figure}

In the SCONE domains, past work has trained listener models with weak supervision (with no intermediate actions between start and end world states) on a subset of the full SCONE training data. 
We use the full training set, and to use a model and training procedure consistent with the SAIL setting, train listener and speaker models using the intermediate actions as supervision as well.\footnote{Since the pragmatic inference procedure we use is agnostic to the models' training method, it could also be applied to the models of \newcite{guu2017language}; however we find that pragmatic inference can improve even upon our stronger base listener models.}
The evaluation method and test data are the same as in past work on SCONE: models are provided with an initial world state and a sequence of 5 instructions to carry out, and are evaluated on their accuracy in reaching the intended final world state. 

Results are reported in \autoref{tbl:scone_follower_results}. 
We see gains from the rational system $L_0 \cdot L_1$ in both the Alchemy and Scene domains. 
The pragmatic inference procedure allows correcting errors or overly-literal interpretations from the base listener. An example is shown in \autoref{fig:alchemy_listener}. The base listener (left) interprets \emph{then to orange's other side} incorrectly, while the rational listener discounts this interpretation (it could, for example, be better described by \emph{to the left of blue}) and produces the action the descriptions were meant to describe (right). To the extent that human annotators already account for pragmatic effects when generating instructions, examples like these suggest that our model's explicit reasoning is able to capture interpretation behavior that the base sequence-to-sequence listener model is unable to model.

\subsection{Instruction generation}
\label{sec:generation_methodology}
\begin{table}[t]
\centering
\footnotesize
\scalebox{0.94}{
\hspace{-1.5mm}
\begin{tabular}{rcccc}
  \toprule
  speaker & SAIL & Alchemy & Scene & Tangrams \\
  \midrule
  DBW & \phantom{+}70.9 &  \phantom{+}--- & \phantom{+}--- & \phantom{+}--- \\
  \hdashline \\[-0.8em]
  $S_0$ & \phantom{+}62.8 &  \phantom{+}29.3 & \phantom{+}31.3  & \phantom{+}60.0 \\
  $S_0 \cdot S_1$ & \phantom{+}\bf 75.2 &  \phantom{+} \bf75.3 & \phantom{+} \bf69.3 & \phantom{+}\bf88.0 \\
  \emph{accuracy gain} & +12.4 & +46.0 & +38.0 & +28.0 \\
  \hdashline \\[-0.8em]
  human-generated & \phantom{+}73.2 &  \phantom{+}83.3 & \phantom{+}78.0 & \phantom{+}66.0 \\
  \bottomrule
\end{tabular}
}
\caption{\label{tbl:speaker_human_evals} Instruction generation results. We report the accuracies of human evaluators at following the outputs of the speaker systems (as well as other humans) on 50-instance samples from the SAIL dataset and SCONE domains. DBW is the system of \newcite{daniele2016navigational}. Bold numbers are new state-of-the-art results.
}
\end{table}

\begin{table}[t]
\centering
\footnotesize
\begin{tabular}{rcccc}
  \toprule
speaker & SAIL & Alchemy & Scene & Tangrams \\
  \midrule
DBW & 11.00 & --- & --- & --- \\
  \hdashline \\[-0.8em]
  $S_0$ & 12.04 &  19.34 & 18.09 & 21.75 \\
  $S_0 \cdot S_1$ & 10.78 &  18.70 & 27.15 & 23.03 \\ 
\emph{BLEU gain} & -1.26 &  -0.64 &  +9.06 & +1.28 \\
  \hdashline \\[-0.8em]
\emph{accuracy gain}  & \multirow{2}{*}{+12.4} &\multirow{2}{*}{+46.0} &\multirow{2}{*}{+38.0} & \multirow{2}{*}{+28.0} \\
(from \autoref{tbl:speaker_human_evals}) \\
\bottomrule
\end{tabular}
\caption{\label{tbl:speaker_bleu_evals}
Gains in how easy the directions are to follow are not always associated with a gain in BLEU.
This table shows corpus-level 4-gram BLEU comparing outputs of the speaker systems to human-produced directions on the SAIL dataset and SCONE domains, compared to gains in accuracy when asking humans to carry out a sample of the systems' directions (see \autoref{tbl:speaker_human_evals}).}
\vspace{-1em}
\end{table}
As our primary evaluation for the instruction generation task, we had Mechanical Turk workers carry out directions produced by the speaker models (and by other humans) in a simulated version of each domain. For SAIL, we use the simulator released by \newcite{daniele2016navigational} which was used in their human evaluation results, and we construct simulators for the three SCONE domains. In all settings, we take a sample of 50 action sequences from the domain's test set (using the same sample as \newcite{daniele2016navigational} for SAIL), and have three separate Turk workers attempt to follow the systems' directions for the action sequence.

\autoref{tbl:speaker_human_evals} gives the average accuracy of subjects in reaching the intended final world state across all sampled test instances, for each domain. The ``human-generated'' row reports subjects' accuracy at following the datasets' reference directions. The directions produced by the base speaker $S_0$ are often much harder to follow than those produced by humans (e.g.\ 29.3\% of $S_0$'s directions are correctly interpretable for Alchemy, vs. 83.3\% of human directions). However, we see  substantial gains from the rational speaker $S_0 \cdot S_1$ over $S_0$ in all cases (with absolute gains in accuracy ranging from 12.4\% to 46.0\%), and the average accuracy of humans at following the rational speaker's directions is substantially higher than for human-produced directions in the Tangrams domain.
In the SAIL evaluation, we also include the directions produced by the system of \citet{daniele2016navigational} (DBW), and find that the rational speaker's directions are followable to comparable accuracy. 

We also compare the directions produced by the systems to the reference instructions given by humans in the dataset, using 4-gram BLEU\footnote{See \autoref{sec:appendix} for details on evaluating BLEU in the SAIL setting, where there may be a different number of reference and predicted sentences for a given example.}~\cite{papineni2002bleu} in \autoref{tbl:speaker_bleu_evals}.
Consistent with past work~\cite{krahmer2010empirical}, we find that BLEU score is a poor indicator of whether the directions can be correctly followed. 

\begin{figure}[t]
\centering
\footnotesize
\includegraphics[width=0.55\linewidth]{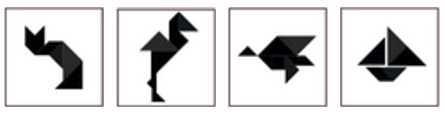}
\includegraphics[width=0.55\linewidth]{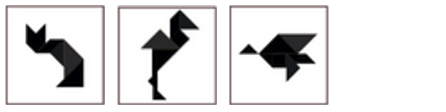}
\includegraphics[width=0.55\linewidth]{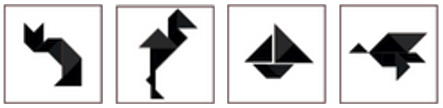}
\\[0.5em]
\begin{tabular}{ll}
\multirow{2}{*}{human} &\texttt{take away the last item} \\
&\texttt{undo the last step} \\ [0.5em]
\multirow{2}{*}{$S_0$} & \texttt{remove the last figure} \\
& \texttt{add it back} \\[0.5em]
\multirow{2}{*}{$S_0\cdot S_1$} & \texttt{remove the last figure} \\
& \texttt{add it back in the 3rd position} \\
\end{tabular}
\caption{\label{fig:tangrams_speaker}Descriptions produced for a partial action sequence  in the Tangrams domain. Neither the human nor base speaker $S_0$ correctly specifies where to add the shape in the second step, while the rational speaker $S_0 \cdot S_1$ does.}
\vspace{-1em}
\end{figure}

Qualitatively, the rational inference procedure is most successful in fixing ambiguities in the base speaker model's descriptions. \autoref{fig:tangrams_speaker} gives a typical example of this for the last few timesteps from a Tangrams instance.
The base speaker correctly describes that the shape should be added back, but does not specify where to add it, which could lead a listener to add it in the same position it was deleted. The human speaker also makes this mistake in their description. This speaks to the difficulty of describing complex actions pragmatically even for humans in the Tangrams domain. The ability of the pragmatic speaker to produce directions that are easier to follow than humans' in this domain (\autoref{tbl:speaker_human_evals}) shows that the pragmatic model can generate something different (and in some cases better) than the training data.

\subsection{Pragmatics as model combination}
Finally, our rational models can be viewed as pragmatically-motivated model combinations, producing candidates using base listener or speaker models and reranking using a combination of scores from both.
We want to verify that a rational listener using $n$ ensembled base listeners and $n$ base speakers outperforms a simple ensemble of $2n$ base listeners (and similarly for the rational speaker).

Fixing the total number of models to 20 in each listener experiment, we find that the rational listener (using an ensemble of 10 base listener models and 10 base speaker models) still substantially outperforms the ensembled base listener (using 20 base listener models): accuracy gains are 68.5 $\to$ 71.6\%, 70.1 $\to$ 72.0\%, 71.9 $\to$ 72.7\%, and 69.1 $\to$ 69.6\% for SAIL single-sentence Rel, Alchemy, Scene, and Tangrams, respectively.

For the speaker experiments, fixing the total number of models to 10 (since inference in the speaker models is more expensive than in the follower models), we find similar gains as well: the rational speaker improves human accuracy at following the generated instructions from 61.9 $\to$ 73.4\%, 30.7 $\to$ 74.7\%, 32.0 $\to$ 66.0\%, 58.7 $\to$ 92.7\%, for SAIL, Alchemy, Scene, and Tangrams, respectively.\footnote{The accuracies for the base speakers are slightly different than in \autoref{tbl:speaker_human_evals}, despite being produced by the same systems, since we reran experiments to control as much as possible for time variation in the pool of Mechanical Turk workers.}

\section{Conclusion}
We have demonstrated that a simple procedure for pragmatic inference, with a unified treatment for speakers and listeners, obtains improvements for instruction following as well as instruction generation in multiple settings. The inference procedure is capable of reasoning about sequential, interdependent actions in non-trivial world contexts.
We find that pragmatics improves upon the performance of the base models for both tasks, in most cases substantially. While this is perhaps unsurprising for the generation task, which has been discussed from a pragmatic perspective in a variety of recent work in NLP, it is encouraging that pragmatic reasoning can also improve performance for a grounded listening task with sequential, structured output spaces. 

\section*{Acknowledgments}

We are grateful to Andrea Daniele for sharing the SAIL simulator and their system's outputs, to Hongyuan Mei for help with the dataset, and to Tom Griffiths and Chris Potts for helpful comments and discussion. This work was supported by DARPA through the Explainable Artificial Intelligence (XAI) program. DF is supported by a Huawei / Berkeley AI fellowship. JA is supported by a Facebook graduate fellowship.

\bibliography{jacob,refs}
\bibliographystyle{acl_natbib}

\include{appendix}

\end{document}

%% file: appendix.tex
\appendix

\section{Supplemental Material}
\label{sec:appendix}

\begin{table}
  \hskip-2.1mm
\footnotesize
\centering
\begin{tabularx}{1.03\linewidth}{Xll}
\toprule
type & arguments & contextual embedding \\
\midrule
\multicolumn{3}{l}{Alchemy} \\
\textsc{Mix} & source $i$ & contents of $i$ \\
\textsc{Pour} & source $i$, target $j$ & contents of $i$ and $j$ \\
\textsc{Drain}& amount $a$, source $i$ & $a$, contents of $i$ \\
\midrule
\multicolumn{3}{l}{Scene} \\
\textsc{Enter} & color $c$, source $i$ & people at $i-1$ and $i+1$ \\
\textsc{Exit} & source $i$ & people at $i$, $i-1$, $i+1$ \\
\textsc{Move} & source $i$, target $j$ & people at $i$, $j-1$, $j+1$ \\
\textsc{Switch} & source $i$, target $j$ & people at $i$ and $j$ \\
\textsc{TakeHat} & source $i$, target $j$ & people at $i$ and $j$ \\
\midrule
\multicolumn{3}{l}{Tangrams} \\
\textsc{Remove} & position $i$ & --- \\
\textsc{Swap} & positions $i$ and $j$ & --- \\
\multirow{2}{*}{\textsc{Insert}} & \multirow{2}{*}{position $i$, shape $s$} & index of step when $s$ \\
& & was removed\\
\bottomrule
\end{tabularx}
\caption{\label{tbl:scone_actions} Action types, arguments, and elements of the world state or action history that are extracted to produce contextual action embeddings.}
\end{table}
\subsection{SCONE listener details}
\label{sec:scone_listener_details}

We factor action production in each of the three SCONE domains, separately predicting the action type and the arguments specific to that action type. Action types and arguments are listed in the first two columns of \autoref{tbl:scone_actions}. 
For example, Alchemy's actions involve predicting the action type, a potential source beaker index $i$ and target beaker index $j$, and potential amount to drain $a$. All factors of the action (the type and options for each argument) are predicted using separate attention mechanisms, which produce a vector $q_f$ giving unnormalized scores for factor $f$ (e.g. scoring each possible type, or each possible choice for the argument).

We also obtain state-specific embeddings of actions, to make it easier for the model to learn relevant features from the state embeddings (e.g. rather than needing to learn to select the region of the state vector corresponding to the 5th beaker in the action \textsc{Mix(5)} in Alchemy, this action's contextual embedding encodes the current content of the 5th beaker). We incorporate these state-specific embeddings into computation of the action probabilities using a bilinear bonus score:
\begin{equation*}
b(a) = q^\top W_{qa} a + w_{a}^\top a
\end{equation*}
where $q$ is the concatenation of all $q_f$ factor scoring vectors, and $W_{qa}$ and $w_a$ are a learned parameter matrix and vector, respectively. This bonus score $b(a)$ for each action is added to the unnormalized score for the corresponding action $a$ (computed by summing the entries of the $q_f$ vectors which correspond to the factored action components), and the normalized output distribution is then produced using a softmax over all valid actions. 

\subsection{SAIL speaker details}
\label{sec:sail_speaker_details}
Since our speaker model operates on segmented action sequences, we train a route segmenter on the training data and then predict segmentations for the test data. This provides a closer comparison to the generation system of \newcite{daniele2016navigational} which did not use segmented routes. The route segmenter runs a bidirectional LSTM over the concatenated state and action embeddings (as in the speaker encoder), then uses a logistic output layer to classify whether the route should be split at each possible timestep.
We also collapse consecutive sequences of forward movement actions into single actions (e.g.\ \textsc{Move4} representing four consecutive forward movements), which we found helped prevent counting errors (such as outputting \emph{move forward three} when the correct route moved forward four steps).

\subsection{SCONE speaker details}
\label{sec:scone_speaker_details}
We use a one-hot representation of the arguments  (see \autoref{tbl:scone_actions}) and contextual embedding (as described in~\ref{sec:scone_listener_details}) for each action $a_t$ as input to the SCONE speaker encoder at time $t$ (along with the representation $e_t$ of the world state, as in SAIL). 
Since SCONE uses a monotonic, one-to-one alignment between actions and direction sentences, the decoder does not use a learned attention mechanism but fixes the contextual representation $z_k$ to be the encoded vector at the action corresponding to the sentence currently being generated.

\subsection{Training details}
\label{sec:training_details}

\begin{table}
\footnotesize
\centering
\begin{tabular}{clccc}
\toprule
& & dropout & hidden & attention \\
model & domain & rate & dim & dim \\
\midrule
$L_0$ & SAIL & 0.25 & 100 & 100 \\
$L_0$ & Alchemy & 0.1 & 50 & 50 \\
$L_0$ & Scene & 0.1 & 100 & 100 \\
$L_0$ & Tangrams & 0.3 & 50 & 100 \\
\midrule
$S_0$ & SAIL & 0.25 & 100 & 100 \\
$S_0$ & Alchemy & 0.3 & 100 & -- \\
$S_0$ & Scene & 0.3 & 100 & -- \\
$S_0$ & Tangrams & 0.3 & 50 & -- \\
\bottomrule
\end{tabular}
\caption{\label{tbl:experiment_parameters}Hyperparameters for the base listener ($L_0$) and speaker ($S_0$) models. The SCONE speakers do not use an attention mechanism.}
\end{table}
We optimize model parameters using \textsc{adam} \cite{kingma2015adam} with default hyperparameters and the initialization scheme of \citet{glorot2010understanding}.
All LSTMs have one layer. The LSTM cell in both the listener and the follower use coupled input and forget gates, and peephole connections to the cell state \cite{greff2016lstm}.
We also apply the LSTM variational dropout scheme of \citet{Gal2016Theoretically}, using the same dropout rate for inputs, outputs, and recurrent connections. 
See \autoref{tbl:experiment_parameters} for hyperparameters.
We perform early stopping using the evaluation metric (accuracy for the listener and BLEU score for the speaker) on the development set. 

\subsection{Computing BLEU for SAIL}
To compute BLEU in the SAIL experiments, as the speaker models may choose produce a different number of sentences for each route than in the true description, we obtain a single sequence of words from a multi-sentence description produced for a route by concatenating the sentences, separated by end-of-sentence tokens.
We then calculate corpus-level 4-gram BLEU between all these sequences in the test set and the true multi-sentence descriptions (concatenated in the same way).